# Understanding Generative AI Use in Programming MOOCs

The Role of Course Context and Learner Characteristics

Marina Lepp
Institute of Computer Science
University of Tartu
Tartu, Estonia
marina.lepp@ut.ee

## ABSTRACT

The increasing availability of generative artificial intelligence (GenAI) tools, such as ChatGPT and code-completion assistants, raises questions about how learners integrate these tools into learning activities, particularly in MOOCs that attract diverse participant populations. This study examines the use of GenAI in two programming MOOCs taught in Estonian that differ in duration, workload, topic complexity, and assignment volume: About Programming (4 weeks, 26 expected hours, n = 187) and Introduction to Programming (8 weeks, 78 expected hours, n = 182). Post-course questionnaire data were analyzed using non-parametric statistical methods to examine self-reported adoption, usage frequency, and purposes of GenAI use across courses and learner backgrounds. The results show that GenAI adoption was widespread in both MOOCs, with no statistically significant differences by gender, age, education level, or prior programming experience. However, participants in the longer, more extensive MOOC reported significantly higher usage frequency and were more likely to use GenAI for debugging and idea generation. Reported usage frequency for code explanation and debugging was also higher in the longer course. Exploratory analyses found limited relationships between GenAI use and learning-related outcomes. The findings suggest that course context may play a greater role than learner characteristics in shaping how GenAI tools are used. These results provide implications for instructional design and guidance in programming education.

**ACM Reference format:**

Marina Lepp. 2027. Understanding Generative AI Use in Programming MOOCs: The Role of Course Context and Learner Characteristics. In *Proceedings of ACM Technical Symposium on Computer Science Education (SIGCSE TS'27), February 17–20, 2027, Sacramento, CA, USA. ACM, New York, NY, USA, 7 pages.*

## 1 Introduction

The increasing adoption of generative artificial intelligence (GenAI) tools, such as ChatGPT and code-generation assistants, is transforming programming education [29]. These tools can generate code, explain concepts, and assist with debugging, offering new opportunities for supporting student learning [3, 16]. As a result, students increasingly use GenAI tools as readily available sources of guidance and feedback [4, 18]. Despite their growing adoption, the educational impact of GenAI tools remains uncertain. Research reports both benefits, such as improved performance and learning efficiency, and risks, including overreliance and reduced cognitive engagement [6, 26]. These mixed findings suggest that the effects of GenAI depend not only on whether students use these tools, but also on how they use them.

Understanding GenAI use across course contexts and learner backgrounds is particularly relevant in massive open online courses (MOOCs), which attract participants with diverse ages, education, and prior experience [22], and may inform curriculum design, learner support, and guidance on the responsible use of GenAI in programming education. This study compares two introductory programming MOOCs that differ in duration, workload, topic complexity, and assignment volume, distinguishing adoption, purpose prevalence, and reported frequency, and examining associations with learner characteristics and recorded performance and activity. It addresses four research questions:

**RQ1:** How do reported GenAI adoption and overall usage frequency differ between different programming MOOCs?
**RQ2:** How do the prevalence and reported frequency of specific purposes of GenAI use differ between programming MOOCs?
**RQ3:** How do reported GenAI adoption and usage frequency vary across gender, age, educational level, and prior programming experience?
**RQ4:** What relationships exist between reported GenAI use and learning-related outcomes?

## 2 Related Work

GenAI tools offer on-demand programming guidance [4, 5, 18], including in MOOC settings [7]. However, their use also carries risks: outputs may be incorrect or misleading [19], and overreliance may hinder cognitive engagement [17, 26]. Research therefore examines the conditions under which these tools support learning [6].

Empirical findings regarding the relationship between GenAI use and learning outcomes are mixed. Some studies report improvements in student performance, self-efficacy, and computational thinking when AI tools are used appropriately [1, 9]. However, other studies find no significant differences [35, 37] or even negative effects, particularly when students rely heavily on AI-generated solutions without engaging in deeper reasoning [12, 13, 18]. These inconsistencies are often explained by differences in how students use GenAI tools. For example, students who use AI for explanation and conceptual understanding tend to benefit more than those who primarily use it to generate complete solutions [17, 30].

Recent research emphasizes that GenAI use is not uniform but involves diverse interaction patterns. Students engage with AI tools for various purposes, including debugging, idea generation, explanation-seeking, and solution generation [28, 31]. Although GenAI use is widespread, it is often occasional and primarily aimed at understanding concepts, saving time, and improving academic performance [24]. These interaction patterns vary in depth of cognitive engagement, ranging from iterative refinement and critical evaluation to copying generated solutions without modification [2, 27]. Heavy reliance on AI may contribute to superficial learning and an "illusion of competence" [26, 27].

Importantly, GenAI interaction patterns are influenced by both learner characteristics and contextual factors. Learners' habits and motivation shape how they use GenAI tools [33], while fields of study affect both attitudes toward and use of AI chatbots in education [23, 32]. Usage also tends to increase with task complexity and course level [16, 28]. More experienced learners often use AI for refinement and exploration, whereas novices may rely on it to overcome basic difficulties or complete tasks quickly [15, 26]. Recent research further suggests that the effectiveness of GenAI tools depends on the interaction between learner characteristics and the tools' affordances [36].

Evidence about demographic differences is mixed: some studies report gender differences in use or attitudes [25, 32], whereas others find limited differences [23, 34]. Academic level appears to have a limited effect [32], although earlier-stage students may perceive greater GenAI benefits than more advanced students [23]. Higher educational attainment has been linked to more positive AI attitudes [11], but other studies find that education and age do not significantly predict AI perceptions [14]. In contrast, younger individuals may hold more positive AI attitudes [10]. Together, these findings highlight the complexity of how individual and contextual factors shape GenAI use. This motivates distinguishing adoption, purpose prevalence, and reported frequency when comparing diverse learners across programming MOOCs.

## 3 Methods

### 3.1 Context

This study investigates self-reported GenAI use in two programming MOOCs taught in Estonian and using Python: About Programming (MOOC1) and Introduction to Programming (MOOC2). Both courses introduce programming to learners with little or no prior programming experience, but differ in duration, expected workload, content coverage, and assignment volume. About Programming is a shorter introductory course (4 weeks, 26 hours of expected workload), whereas Introduction to Programming is longer and more extensive (8 weeks, 78 hours). About Programming [20] introduces fundamental programming concepts, including variables, data types, conditional statements, loops, functions, regular expressions, data exchange, and file handling. Each week includes two programming assignments and a weekly test. Introduction to Programming [21] provides a more in-depth treatment of introductory programming topics while developing algorithmic thinking skills. In addition to revisiting core concepts, it introduces topics such as lists, user interfaces, and basic graphics programming. Each week includes four programming assignments and a weekly test. GenAI guidance and assessment rules were the same in both MOOCs. Because these course characteristics vary together, the comparison does not isolate the contribution of any individual characteristic.

### 3.2 Sample and Data Collection

The sample comprised participants in the two aforementioned MOOCs during the 2025/2026 academic year. Participants were registered through the university's Study Information System, which collected demographic data, including gender and age. A total of 226 participants (mean age 38.2 years, 72.6% female) were enrolled in About Programming. In Introduction to Programming, 230 participants were enrolled, with a mean age of 37.5 years; 70.9% were female and 29.1% male.

In both MOOCs after registration, all participants received a questionnaire link, which collected additional background data, including educational level and previous programming experience. At the end of the course, participants were invited to complete a post-questionnaire. A total of 187 participants (82.7% of those who registered, mean age 37.7 years, 73.3% female) in About Programming and 182 participants (79.1% of those who registered, mean age 37.4 years, 70.9% female) in Introduction to Programming responded to the post-questionnaire. These respondents constitute the final sample used in the analysis. Comparisons of respondents and nonrespondents did not detect statistically significant differences in age, gender, educational level, or prior programming experience after Holm adjustment (all adjusted $p \geq 0.423$).

The post-course questionnaire assessed participants' self-reported use of GenAI tools during the course. Participants first indicated whether they had used GenAI tools at least once (yes/no). Those who answered yes reported their overall frequency of use using five ordered response categories ranging from "once" to "weekly." GenAI users also indicated which tools they had used, with multiple responses permitted. They also indicated their purposes of use through a multiple-response question with the following options: debugging, idea generation, code explanation, generating code solutions, and answering weekly tests. For each selected purpose, participants reported their frequency of use using the same five response categories. Frequency responses were treated as ordinal variables. For selected analyses, binary variables were created (e.g., whether a participant used GenAI for a specific purpose). Frequency values reported for overall usage include non-users (coded as zero), while analyses among GenAI users are reported separately. Purpose prevalence was analyzed among GenAI users, whereas purpose-specific frequency analyses included only respondents reporting the corresponding purpose. Percentages and statistical analyses reported are based on valid responses for each item, excluding missing data.

Questionnaire participation was voluntary and unrelated to course completion. Participants received information about the study and data use and were asked to provide informed consent. The University of Tartu Research Ethics Committee approved the study (341/T-2).

In addition to questionnaire responses, course performance indicators were obtained from the learning management system. These indicators included each participant's mean weekly test score (maximum 10 points per test, using the highest-scoring attempt for each test), mean number of submission attempts per assignment, and numbers of completed choice and optional exercises. In MOOC2, participants were required to complete at least 8 of the 22 available choice exercises. There were 6 optional exercises in MOOC1 and 7 in MOOC2. All assignments and weekly tests were completed at home without proctoring. Participants were advised to complete the work independently. The final course result was not analyzed separately because it was recorded as a pass/fail outcome and was derived from the same assessment components examined in this study, providing limited additional information. Quiz scores and submission attempts have also been used to examine assessment performance in earlier research on programming MOOCs [8].

## 3.3 Data Analysis

Analyses were performed in Python using NumPy, SciPy, and statsmodels. Descriptive statistics summarized GenAI adoption, reported frequency, and purposes. Non-parametric tests were selected to analyze differences because several variables were ordinal or did not meet the assumptions of normality. Between-course comparisons used Mann–Whitney U tests for frequency and Pearson's chi-square tests ($\chi^2$) for binary outcomes. P-values were obtained through restricted permutations. Within each course, adoption differences across gender, age group, educational level, and prior programming experience were examined using chi-square tests. Frequency among GenAI users was compared using Mann–Whitney U tests for two groups and Kruskal–Wallis H tests for more than two groups. Learning-related outcomes were compared between users and non-users using Mann–Whitney U tests and related to overall usage frequency using Spearman correlations with permutation-based p-values. Effect sizes included Cramér's V, rank-biserial correlations, rank epsilon-squared, and Spearman coefficients. Pointwise 95% bootstrap confidence intervals were calculated for rank-biserial and Spearman correlations. Holm correction was applied separately to overall course comparisons, purpose-specific comparisons, learner-characteristic comparisons, tool comparisons, non-response comparisons, and learning outcomes analyses within each course. Missing values were excluded from the relevant analyses. Tests were two-sided where applicable. Statistical significance was evaluated at adjusted $p < 0.05$.

# 4 Results

## 4.1 GenAI Use Across Different MOOCs

GenAI tools were widely used in both MOOCs: 142 of 187 respondents in MOOC1 (75.9%) and 150 of 182 respondents in MOOC2 (82.4%) reported using them at least once. The between-course comparison did not detect a statistically significant difference in adoption ($\chi^2 = 2.35$, Cramér's V = 0.080, $p = 0.113$).

However, differences emerged in reported usage frequency between the MOOCs. Although the median frequency of GenAI use was 2 in both MOOCs, participants in MOOC2 reported significantly higher overall usage frequency ($U = 13{,}243$, $r = 0.222$, 95% CI [0.125, 0.319], $p < 0.001$). The mean frequency in MOOC1 was 1.57 (SD = 1.27), whereas in MOOC2 it was 2.08 (SD = 1.40). In particular, a larger proportion of MOOC2 participants reported using GenAI weekly or almost weekly (18.7%) compared to MOOC1 participants (10.2%) (Figure 1).

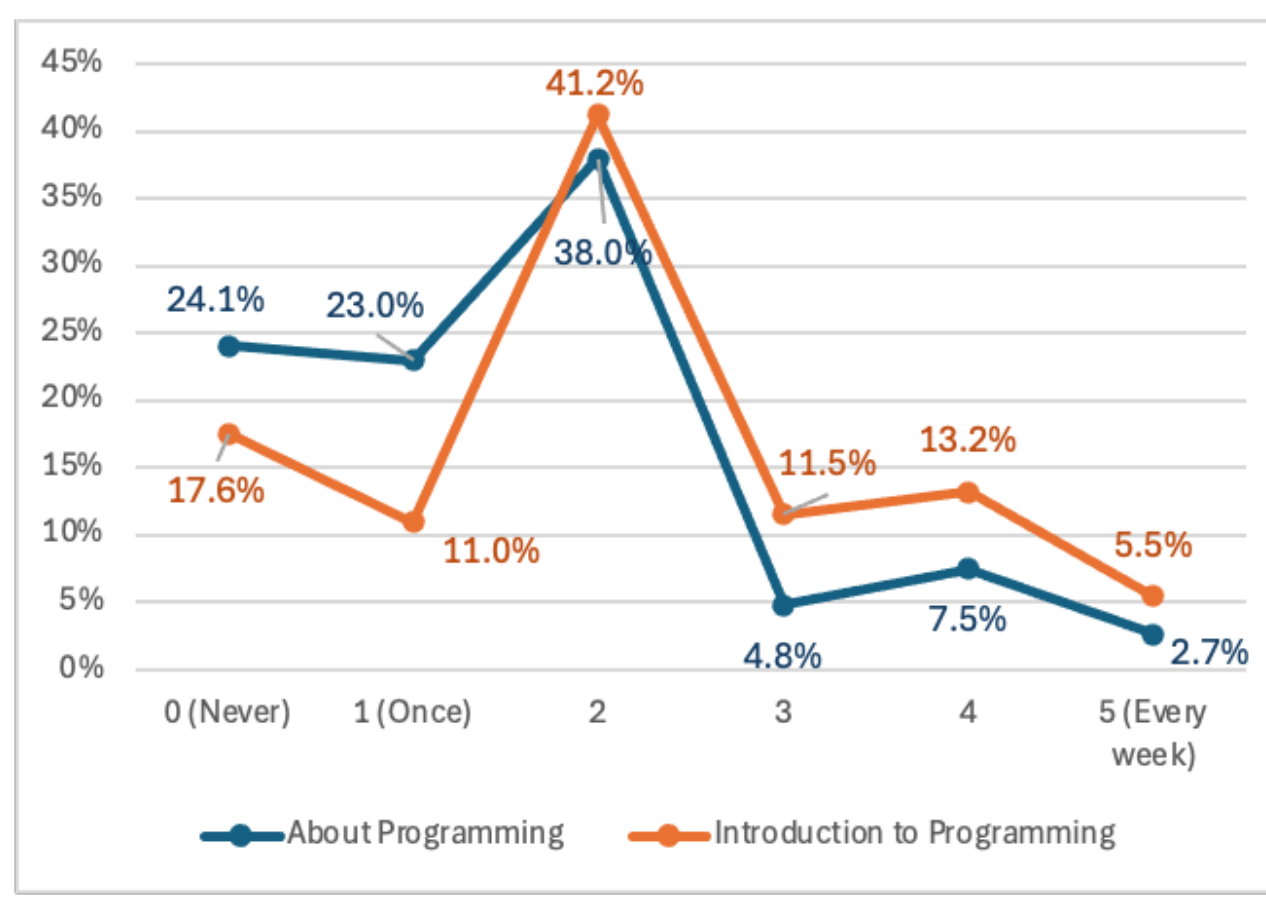

**Figure 1: GenAI usage frequency across MOOCs**

The frequency difference remained significant among GenAI users ($U = 7{,}979$, $r = 0.251$, 95% CI [0.140, 0.361], $p < 0.001$). Among these users, the mean frequency of use was again higher in MOOC2 (M = 2.53, SD = 1.12) than in MOOC1 (M = 2.06, SD = 1.04).

When participants reported the GenAI tools they used, ChatGPT was by far the most frequently mentioned tool in both courses (88.7% in MOOC1 and 82.0% in MOOC2). Bing/M365 Copilot was reported more often in MOOC2 than in MOOC1 (14.7% vs. 7.0%; $\chi^2 = 4.35$, V = 0.122, p = 0.037), while Gemini showed a moderate level of use in both courses (10.6% and 12.0%, respectively). Other tools, including Claude, Perplexity, DeepSeek, and Grok, were reported only by a small proportion of participants.

## 4.2 Purposes of GenAI Use

Significant differences were observed among GenAI users in how they used GenAI tools across MOOCs. Participants in MOOC2 were more likely to report using GenAI for debugging code and for idea generation (Table 1). No statistically significant differences were found for code explanation, obtaining code solutions, or answering weekly tests.

**Table 1: Purposes of GenAI use by MOOC**

| Use-case | MOOC1 n (%) | MOOC2 n (%) | $\chi^2$ | V | p |
|---|---|---|---|---|---|
| Debugging | 108 (76.1%) | 133 (88.7%) | 8.05 | 0.166 | 0.023 |
| Idea generation | 38 (26.8%) | 63 (42.0%) | 7.49 | 0.160 | 0.040 |
| Code explanation | 69 (48.6%) | 65 (43.3%) | 0.81 | 0.053 | 1.000 |
| Code solutions | 16 (11.3%) | 18 (12.0%) | 0.04 | 0.011 | 1.000 |
| Answering tests | 4 (2.8%) | 8 (5.3%) | 1.17 | 0.063 | 1.000 |

Differences were also observed in the frequency of GenAI use across specific use cases (Table 2). Debugging and code explanation were reported more frequently in MOOC2. The frequency differences for idea generation, code generation, and answering weekly tests were not statistically significant after correction, although the mean values were higher in MOOC2 for all use cases.

**Table 2: Frequency of GenAI use by use case across MOOCs**

| Use-case | MOOC1 M (SD) | MOOC2 M (SD) | U | r [95% CI] | p |
|---|---|---|---|---|---|
| Debugging | 2.10 (0.98) | 2.64 (1.12) | 5147.5 | 0.283 [0.162, 0.403] | <0.001 |
| Idea generation | 2.11 (1.13) | 2.37 (1.13) | 1018 | 0.150 [-0.062, 0.354] | 1.000 |
| Code explanation | 2.30 (1.08) | 3.08 (1.16) | 1399 | 0.376 [0.213, 0.537] | <0.001 |
| Code solutions | 2.25 (1.29) | 3.28 (1.02) | 73.5 | 0.490 [0.139, 0.792] | 0.090 |
| Answering tests | 2.25 (1.89) | 3.13 (1.36) | 10.5 | 0.344 [-0.375, 0.969] | 1.000 |

## 4.3 Differences by Learner Characteristics

Within each course, the analyses did not detect statistically significant differences in GenAI adoption or reported frequency among users across gender (Table 3), age group (Table 4), educational level (Table 5), or prior programming experience (Table 6). All sixteen Holm-adjusted p-values were 1.000 and are therefore omitted from Tables 3–6.

**Table 3: GenAI use and frequency of use by gender**

| Course | Female | Male | | |
|---|---|---|---|---|
| | n (%) | n (%) | $\chi^2$ | V |
| MOOC1 | 105 (76.6%) | 37 (74.0%) | 0.14 | 0.027 |
| MOOC2 | 104 (80.6%) | 46 (86.8%) | 0.99 | 0.074 |
| | Frequency M (SD) | | U | r [95% CI] |
| MOOC1 | 2.11 (1.06) | 1.92 (0.98) | 2139 | 0.101 [-0.105, 0.303] |
| MOOC2 | 2.55 (1.16) | 2.48 (1.01) | 2430.5 | 0.016 [-0.162, 0.195] |

**Table 4: GenAI use and frequency of use by age**

| Course | <=24 | 25-34 | 35-44 | 45-54 | >=55 | | |
|---|---|---|---|---|---|---|---|
| | n (%) | n (%) | n (%) | n (%) | n (%) | $\chi^2$ | V |
| MOOC1 | 10 (83.3%) | 46 (73.0%) | 51 (77.3%) | 27 (75.0%) | 8 (80.0%) | 0.83 | 0.066 |
| MOOC2 | 14 (70.0%) | 54 (84.4%) | 47 (87.0%) | 22 (78.6%) | 13 (81.3%) | 3.39 | 0.137 |
| | M (SD) | M (SD) | M (SD) | M (SD) | M (SD) | H | $\varepsilon^2$ |
| MOOC1 | 2.20 (1.40) | 2.04 (1.07) | 2.10 (0.96) | 2.00 (0.96) | 2.00 (1.31) | 0.74 | 0.005 |
| MOOC2 | 2.36 (1.08) | 2.44 (1.02) | 2.51 (1.16) | 2.82 (1.18) | 2.62 (1.33) | 2.06 | 0.014 |

**Table 5: GenAI use by education level**

| Course | Basic / Secondary | Bachelor / Applied | Master or higher | | |
|---|---|---|---|---|---|
| | n (%) | n (%) | n (%) | $\chi^2$ | V |
| MOOC1 | 49 (73.1%) | 41 (77.4%) | 52 (77.6%) | 0.45 | 0.049 |
| MOOC2 | 57 (82.6%) | 45 (81.8%) | 48 (82.8%) | 0.02 | 0.010 |
| | M (SD) | M (SD) | M (SD) | H | $\varepsilon^2$ |
| MOOC1 | 2.00 (0.98) | 2.24 (1.09) | 1.98 (1.06) | 2.11 | 0.015 |
| MOOC2 | 2.42 (1.08) | 2.58 (1.12) | 2.60 (1.15) | 1.00 | 0.007 |

**Table 6: GenAI use and frequency of use by previous experience in programming**

| Course | With experience | Without experience | | |
|---|---|---|---|---|
| | n (%) | n (%) | $\chi^2$ | V |
| MOOC1 | 45 (70.3%) | 85 (77.3%) | 1.04 | 0.077 |
| MOOC2 | 86 (85.1%) | 44 (80%) | 0.68 | 0.066 |
| | Frequency M (SD) | | U | r [95% CI] |
| MOOC1 | 2.18 (1.11) | 1.99 (0.93) | 2076.5 | 0.086 [-0.108, 0.276] |
| MOOC2 | 2.52 (1.11) | 2.48 (1.17) | 1964.5 | 0.038 [-0.157, 0.232] |

## 4.4 GenAI Use and Learning Outcomes

In MOOC1, no significant differences were observed in weekly test scores or assignment submissions (Table 7), although non-users completed significantly more optional exercises than GenAI users (M = 2.09 versus 0.96; r = −0.251, 95% CI [−0.432, −0.067], adjusted p = 0.027). In MOOC2, no significant differences were found across any performance indicators; however, non-users completed slightly more choice exercises, although the difference did not reach statistical significance after correction. Correlations between overall GenAI usage frequency and performance indicators in both courses ranged from −0.139 to 0.117; none was statistically significant after Holm correction (all adjusted p ≥ 0.289).

**Table 7: GenAI use and performance in MOOCs**

| | AI users M (SD) | Non-users M (SD) | U | r | p |
|---|---|---|---|---|---|
| **MOOC1** | | | | | |
| Test score | 9.79 (0.25) | 9.86 (0.16) | 2793.5 | -0.126 | 0.519 |
| Submissions | 1.70 (1.22) | 1.60 (0.64) | 2910 | -0.089 | 0.620 |
| Optional tasks | 0.96 (1.60) | 2.09 (2.39) | 2392.5 | -0.251 | 0.027 |
| **MOOC2** | | | | | |
| Test score | 9.89 (0.17) | 9.88 (0.14) | 2655.5 | 0.106 | 1.000 |
| Submissions | 1.39 (0.35) | 1.46 (0.40) | 2096.5 | -0.126 | 1.000 |
| Choice tasks | 10.38 (4.25) | 12.06 (5.03) | 1886.5 | -0.214 | 0.428 |
| Optional tasks | 0.98 (1.73) | 1.16 (1.89) | 2299 | -0.042 | 1.000 |

## 5 Discussion

This study examined self-reported GenAI use in two programming MOOCs with different course characteristics. It distinguished overall adoption, the prevalence of specific purposes, and reported usage frequency, while examining learner characteristics and associations with recorded assessment performance and course activity.

First, the results show that GenAI adoption was widespread among respondents in both courses, with no statistically significant difference detected in the proportion of participants who reported using these tools. This aligns with prior research indicating that GenAI tools are becoming a common part of learning practices [16, 28]. At the same time, the overall pattern of usage observed in this study is consistent with large-scale evidence indicating that students tend to use GenAI tools widely but often only occasionally, primarily to support understanding and complete learning tasks efficiently [24].

However, differences emerged in reported usage frequency. Participants in the longer, more extensive course (Introduction to Programming) reported significantly higher overall usage frequency. Importantly, this difference remained significant even when considering only participants who used GenAI tools, showing that higher frequency was also evident within the user groups. The estimated overall frequency differences were modest. This finding is consistent with prior research suggesting that GenAI use may increase with task complexity and course level, as learners turn to AI tools more frequently when facing more difficult tasks [16, 28]. One possible explanation is that greater problem-solving demands in more extensive courses encourage learners to seek external support more frequently. A longer duration and greater assignment volume may also provide more opportunities for GenAI use. These possibilities are consistent with previous descriptions of GenAI tools as on-demand sources of guidance and feedback [4, 5, 18]. However, the present study did not measure problem-solving effort or establish whether GenAI functioned as effective scaffolding.

Second, the purpose-specific findings demonstrate why the prevalence of a purpose and its frequency among respondents reporting that purpose should be considered separately. Debugging and idea generation were more prevalent among GenAI users in the longer MOOC, whereas debugging and code explanation were reported more frequently among respondents using GenAI for those purposes. Code explanation illustrates the distinction particularly clearly: evidence of higher frequency among its users was not accompanied by evidence of higher prevalence. Reporting only whether participants used GenAI for an activity would therefore have missed part of the observed pattern. The estimated frequency difference for required-assignment code solutions also favored the longer MOOC but was imprecise and did not meet the significance threshold after Holm correction. This remains an uncertain finding rather than an established between-course difference.

These patterns complement previous accounts of diverse GenAI activities in programming education [28, 31]. However, a reported purpose does not establish the quality of the interaction. For example, using GenAI for debugging could involve critically evaluating an explanation, testing a suggested correction, or accepting generated code without further examination. Previous research distinguishes such interaction patterns [2, 27], but they were not directly observed in this study. Higher reported frequency should therefore not be interpreted as evidence of deeper understanding, iterative refinement, or more effective learning.

Third, no statistically significant differences in GenAI adoption or reported usage frequency were found across participant characteristics, including gender, age, educational level, and prior programming experience. The estimated associations were generally small, although the results do not establish equivalence between learner groups. These findings contribute evidence from a heterogeneous programming MOOC population to a literature reporting mixed demographic patterns in GenAI use, attitudes, and perceived benefits [23, 25, 32, 34]. One possible explanation is that ease of access to GenAI tools may reduce some barriers to adoption across learner groups. However, perceived ease of use and technical barriers were not measured, so this explanation remains tentative. Although the findings originate from two Estonian-language MOOCs, they may provide insights relevant to other online programming-learning contexts that attract heterogeneous learner populations. However, the present findings should be understood specifically as evidence about reported adoption and frequency. In particular, limited differences by prior programming experience do not imply that novice and experienced learners evaluate or apply generated outputs in the same ways.

Fourth, exploratory analyses revealed limited evidence of associations between GenAI use and recorded assessment performance and course activity. With the exception of optional exercises in the shorter course, GenAI users and non-users demonstrated similar levels of performance and engagement across the measured indicators. Furthermore, more frequent GenAI use was not associated with better performance in either course. These findings suggest that while GenAI use varies across course contexts, greater use does not necessarily translate into measurable differences in achievement or engagement. This observation is consistent with prior research reporting mixed relationships between GenAI use and learning outcomes [1, 9, 13, 35, 37]. Earlier research on the About Programming MOOC found that more active use of support resources among completers was associated with more submission attempts and lower quiz scores [8]. Although that study did not examine GenAI, it raises the possibility that greater use of assistance may reflect learners' difficulties or support needs. Previous research also emphasizes that learning outcomes may depend on how learners integrate GenAI into their learning processes, not only on how often they use it [17, 30]. This offers a possible explanation for the present pattern, but those learning processes were not examined directly.

Taken together, the findings suggest that GenAI use is context dependent. In particular, course characteristics, such as duration, complexity, and task demands, may help shape how frequently and for what purposes GenAI tools are used. Recent research emphasizes interactions between learner characteristics and GenAI affordances in supporting self-regulated learning [36]. The present findings complement that perspective by identifying course-context differences that warrant further investigation. However, because topic complexity, workload, assignment volume, and course duration varied simultaneously between the two MOOCs, the present study cannot determine the relative contribution of each factor to the observed differences.

## 6 Conclusion

This study examined self-reported GenAI use in two programming MOOCs with different course characteristics. GenAI adoption was widespread among respondents in both courses, with no statistically significant between-course difference detected. However, participants in the longer, more extensive course reported higher overall usage frequency. Among GenAI users, debugging and idea generation were more prevalent in that course, whereas debugging and code explanation were reported more frequently among respondents using GenAI for those purposes. No statistically significant differences in adoption or reported frequency were detected across gender, age, educational level, or prior programming experience. Exploratory analyses provided limited evidence of associations with recorded assessment performance and course activity, although non-users completed more optional exercises in the shorter course. Overall, the findings highlight the value of distinguishing adoption, purpose prevalence, and reported frequency when examining GenAI use. They contribute evidence from authentic programming-learning environments and may inform instructional guidance for diverse learner populations.

The findings have several implications for programming education. First, widespread reported use suggests that educators should acknowledge the presence of GenAI tools and clarify expectations for their use in learning activities and assessments. Guidance could address commonly reported activities, such as debugging and code explanation, while encouraging students to critically evaluate generated outputs and reflect on their problem-solving processes. Second, the differences in reported use between courses suggest that course characteristics should be considered when planning such guidance. Although common guidance could be offered across learner groups, the absence of statistically significant demographic differences does not establish that the same support would be equally effective for everyone. The effectiveness of different instructional approaches therefore remains a question for further research.

This study has several limitations. First, GenAI use was assessed through self-report, which may be subject to recall bias or inaccuracies in reporting. Despite the relatively high questionnaire response rates, respondents may differ from nonrespondents in characteristics associated with course participation or GenAI use. Second, the study focuses on two MOOCs in a specific context, which may limit generalizability to other courses or educational settings. Third, although the study included recorded assessment performance and course activity, it did not examine actual interactions with GenAI tools and could not determine how specific interaction patterns related to learning. Assessments were unproctored, so scores reflect performance under the course conditions rather than verified unaided programming competence. Finally, topic complexity, workload, assignment volume, and course duration varied simultaneously between the MOOCs, preventing their individual contributions from being separated.

Future research should combine self-reported data with behavioral data, such as logs of interactions with GenAI tools, to better understand how students use these tools in practice. Direct assessments of learning would help distinguish course performance from the development of programming competence. Additionally, longitudinal studies could examine how usage patterns evolve as learners gain experience. Further research is also needed to explore how instructional interventions can promote effective, critical, and responsible use of GenAI in programming education.

## ACKNOWLEDGMENTS

This work was supported by the Estonian Research Council grant "Developing human-centric digital solutions" (TEM-TA120).